\documentclass{ecai}

\usepackage{latexsym}
\usepackage{amssymb}
\usepackage{amsmath}
\usepackage{amsthm}
\usepackage{booktabs}
\usepackage{enumitem}
\usepackage{graphicx}
\usepackage{svg}
\usepackage{color}
\usepackage{algorithm}
\usepackage{algorithmicx}
\usepackage{algpseudocode}

\newcommand{\BibTeX}{B\kern-.05em{\sc i\kern-.025em b}\kern-.08em\TeX}
\begin{document}
%%%%%%%%%%%%%%%%%%%%%%%%%%%%%%%%%%%%%%%%%%%%%%%%%%%%%%%%%%%%%%%%%%%%%%%%
\begin{frontmatter}

%%% Use this command to specify your submission number.
\paperid{29}

%%% Use this command to specify the title of your paper.
\title{HyDRA: A Hybrid-Driven Reasoning Architecture for Verifiable Knowledge Graphs}

%%% Use this combinations of commands to specify all authors of your paper.
\author[A]{\fnms{Adrian}~\snm{Kaiser}\thanks{Corresponding Author. Email: adrian.kaiser@extensity.ai.}}
\author[A]{\fnms{Claudiu}~\snm{Leoveanu-Condrei}}
\author[A]{\fnms{Ryan}~\snm{Gold}}
\author[A,B]{\fnms{Marius-Constantin}~\snm{Dinu}}
\author[A]{\fnms{Markus}~\snm{Hofmarcher}}

\address[A]{ExtensityAI, Austria}
\address[B]{WU Executive Academy, Austria}

%%% Use this environment to include an abstract of your paper.
\begin{abstract}
The synergy between symbolic knowledge, often represented by Knowledge Graphs (KGs), and the generative capabilities of neural networks is central to advancing neurosymbolic AI. A primary bottleneck in realizing this potential is the difficulty of automating KG construction, which faces challenges related to output reliability, consistency, and verifiability. These issues can manifest as structural inconsistencies within the generated graphs, such as the formation of disconnected \textit{isolated islands} of data or the inaccurate conflation of abstract classes with specific instances. To address these challenges, we propose HyDRA, a \textbf{Hy}brid-\textbf{D}riven \textbf{R}easoning \textbf{A}rchitecture designed for verifiable KG automation. 
Given a domain or an initial set of documents, HyDRA first constructs an ontology via a panel of collaborative neurosymbolic agents. These agents collaboratively agree on a set of competency questions (CQs) that define the scope and requirements the ontology must be able to answer. Given these CQs, we build an ontology graph that subsequently guides the automated extraction of triplets for KG generation from arbitrary documents.
Inspired by design-by-contracts (DbC) principles, our method leverages verifiable contracts as the primary control mechanism to steer the generative process of Large Language Models (LLMs). 
To verify the output of our approach, we extend beyond standard benchmarks and propose an evaluation framework that assesses the functional correctness of the resulting KG by leveraging symbolic verifications as described by the neurosymbolic AI framework, \textit{SymbolicAI}. 
This work contributes a hybrid-driven architecture for improving the reliability of automated KG construction and the exploration of evaluation methods for measuring the functional integrity of its output. The code is publicly available.
\end{abstract}

\end{frontmatter}
%%%%%%%%%%%%%%%%%%%%%%%%%%%%%%%%%%%%%%%%%%%%%%%%%%%%%%%%%%%%%%%%%%%%%%%%

\section{Introduction}

The advancement of neurosymbolic AI hinges on the effective fusion of sub-symbolic pattern recognition, characteristic of neural networks, and the structured, explicit reasoning enabled by symbolic knowledge representations \citep{marcus2001algebraic,auer2018towards,colelough2025neurosymbolic, odense2024semantic}. As such, Knowledge Graphs (KGs) are a core component of the symbolic approach, offering a powerful formalism for encoding information. The prospect of using Large Language Models (LLMs) to automate KG construction promises to bridge these two worlds \citep{zhong2023comprehensive, carta2023iterative,zhang2024extract}, potentially augmenting how intelligent systems acquire and reason over knowledge \citep{marcus2001algebraic,besold2017neural}.

However, reliably constructing a KG from a given dataset is still an unsolved topic \cite{paulheim2016knowledge,li2024knowledge,bayer2024assessing}.
Unconstrained, LLMs generate KGs with structural and semantic flaws that undermine their utility. Outputs are often inconsistent, containing disconnected subgraphs—or \textit{isolated islands}—that break the KG's conceptual integrity, or conflating abstract classes with concrete instances in ways that violate ontological principles or capture KG semantics \cite{bonatti2019knowledge}.

\begin{figure}[ht]
    \centering
    \includegraphics[width=0.9\columnwidth]{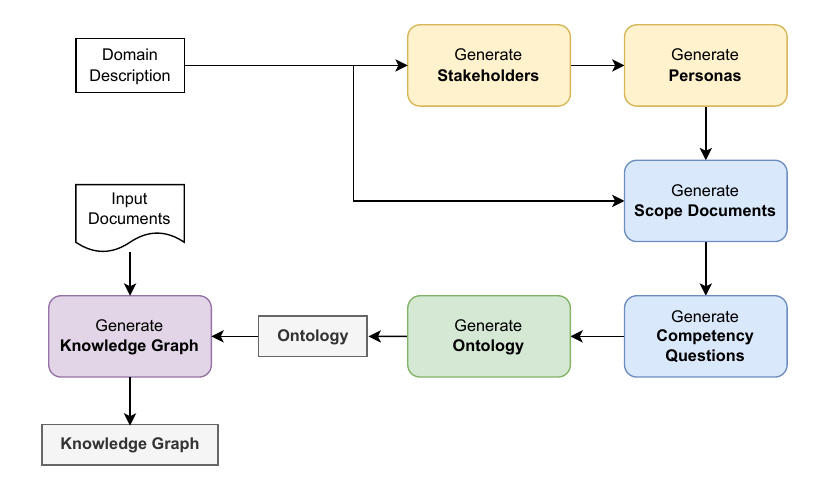}
    \caption{Overview of the HyDRA architecture. The ontology generation process begins with a user-defined domain description. HyDRA first generates stakeholder groups, which define the constraints for generating personas (yellow). It then creates \textit{scope documents} and competency questions (blue), which are used to construct an ontology (green). If input documents are supplied, HyDRA can, as an optional final step, generate a knowledge graph (purple) from a set of documents adhering to the constructed ontology.}
    \label{fig:architecture}
\end{figure}

These issues necessitate time-consuming manual validation and correction, defeating the purpose of automation and creating a significant barrier to deploying trustworthy, KG-augmented AI systems.

One of the core issues with KG construction from unstructured data is the problem of \textit{entity resolution} (ER) \cite{newcombe59entityresolution}. In the context of automated KG generation from unstructured input ER refers to the challenge of deciding when and how to consolidate entities or concepts that may refer to the same real-world object but appear differently in the data. This problem is central to ensuring the consistency and accuracy of the generated KG, especially when the KG is constructed iteratively, i.e. from one set of documents at a time.

To address this, we leverage a recent paradigm shift following neurosymbolic architectures \cite{dinu2024symbolicai,almachot2024designing}: treating the construction of KG not as an unconstrained generative task, but as a verifiable multistep generative process. We introduce HyDRA \footnote{Codebase: \url{https://github.com/ExtensityAI/ontology-hydra}} (a \textbf{Hy}brid-\textbf{D}riven \textbf{R}easoning \textbf{A}rchitecture), a system that operationalizes this idea through the principles of Design-by-Contract (DbC) \cite{meyer1997object} and neurosymbolic architecture \cite{dinu2024symbolicai}. HyDRA orchestrates an LLM-driven pipeline using a series of verifiable contracts, where each stage of construction, from requirements gathering to instance population, is governed by explicit, machine-readable specifications. Our architecture implements a neurosymbolic feedback loop in which the symbolic layer (the contract) actively steers and corrects the generative output of the neural layer (the LLM).

The process begins by defining a \textit{scope contract} through the generation of competency questions (CQ), which then serve as formal preconditions for creating the \textit{ontology schema}. This schema, governed by its own \textit{schema contract}, becomes a precondition for the final KG population. This creates an end-to-end contract lineage, ensuring traceability from high-level requirements to low-level data. If a contract's postcondition is violated at any stage, HyDRA's core innovation—a closed-loop verification and repair mechanism—is triggered, automatically re-prompting the LLM or executing corrective actions until the KG is verifiably compliant.

In order to test the quality of the HyDRA constructed KG, we selected a challenging question-answering benchmark from the medical domain. 
We selected question-answering as the initial benchmark as datasets of this kind are readily available,
Since questions and answers are initially unstructured data itself, we indirectly evaluate the quality of the KG by supplying the KG in the context of an LLM and then query for the answers. 
Surprisingly, unconstrained KG construction (i.e. without the constraint of an ontology) performs better on such benchmarks.
We identify the underlying reason and propose a novel evaluation scheme to evaluate KG quality. 
Our goal is to establish a benchmark that directly measures the quality of the graph structure by using graph databases such as Neo4j and converting Q\&A pairs into queries that can be interpreted by such databases.
%However, this poses a surprising number of technical and conceptual challenges that we did not anticipate 
However, this poses a number of technical and conceptual challenges that we had to address, and we are thus still in the process of implementing this procedure.
%Ultimately, the test of a KG built to satisfy specific requirements is whether it can functionally answer them. We therefore eschew purely static evaluation metrics in favor of a dynamic, functional testbed. 
%The generated KG is loaded into a Neo4j graph database, and the initial CQs are programmatically translated into Cypher queries. The query success rate serves as a direct, quantifiable measure of the KG's utility, closing the loop between specification and evaluation.
%The generated KG serves as a direct, quantifiable measure of the utility of the task, closing the loop between specification and evaluation.
In summary, this work makes the following contributions:
\begin{itemize}
    \item A hybrid-driven architecture, HyDRA, that adapts Design-by-Contract principles to orchestrate and verify LLM-driven KG construction.
    \item A \textbf{ closed loop contract-driven repair mechanism} that automatically enforces structural invariants, such as graph connectivity and type consistency, through iterative refinement.
    \item A concept for an \textbf{evaluation framework} to measure the practical utility of a KG by leveraging neurosymoblic design principles.
\end{itemize}

\section{Related Work}

This section situates HyDRA within the context of prior research, focusing on four key areas.
We associate neurosymbolic approaches with KG methodologies and articulate how HyDRA’s contract-centric architecture presents a novel solution.

\subsection{Automated Knowledge-Graph (ABox) Construction}

Systems for populating KGs with instance data (ABoxes) range from classic NLP pipelines like \textit{FRED} \cite{gangemi2017fred} to modern LLM-driven frameworks like \textit{GraphRAG} \cite{robinson2023graphrag} and \textit{AdaKGC} \cite{zhang2023adakgc}. While these tools demonstrate high accuracy in extracting individual triplets, their validation capabilities are often confined to local constraints, such as checking a relation's arity or a triplet's adherence to a basic type. They typically lack mechanisms to enforce global structural invariants across the entire graph.

HyDRA's primary differentiation lies in its introduction of a \textbf{principled, contract-driven verification loop to ensure global structural integrity}. Using standards like the Shapes Constraint Language (SHACL) \cite{shacl2017} or graph-based queries, HyDRA defines postcondition contracts that check for global properties, such as the absence of isolated components or the prevention of class-instance conflation. Its iterative repair mechanism systematically identifies and rectifies these structural violations, producing a graph that is not only rich in facts but also structurally sound.

\subsection{LLMs for Ontology (TBox) Generation}

The automated generation of ontologies (TBoxes) with LLMs has shown significant promise. Frameworks like \textit{LLMs4OL} \cite{funk2023llms4ol} and \textit{OLLM} \cite{baldwin2024ollm} demonstrate the ability of LLMs to generate taxonomic and relational structures from text using prompting and fine-tuning. These methods excel at generating candidate schema components with high recall. The common limitation, however, lies in their validation strategies, which are often separated from the generation process. Validation frequently relies on after-the-fact checks, heuristic filtering, or indispensable human-in-the-loop curation to ensure quality and correctness.

HyDRA addresses this gap by tightly \textbf{coupling generation with in-loop verification}. It introduces machine-checkable contracts—such as ensuring a schema is valid OWL-DL or that all classes possess natural language descriptions—as formal postconditions of the generation step. A contract violation is not merely a terminal failure but a signal that triggers a corrective re-prompting or refinement action. This integration moves ontology generation away from a \textit{generate-then-validate} model towards a more reliable, correct-by-construction approach.

\subsection{Competency-Question (CQ) Engineering}

Since their formalization by Gruninger and Fox \cite{gruninger1995role}, Competency Questions (CQs) have been central to defining an ontology's scope. The manual effort of creating CQs is now being alleviated by LLM-based automation, with techniques like Retrieval-Augmented Generation (RAG) for domain grounding \cite{pan2024retrieval} and multi-persona prompting for diverse perspectives \cite{wu2024multi}. While these methods improve the quality and coverage of CQs, they still treat them primarily as informal, human-readable documentation that guides the subsequent development process.

HyDRA fundamentally changes the role of CQs by \textbf{elevating them to formal, machine-verifiable preconditions}. This is a two-fold innovation. First, the set of CQs becomes the root \textit{scope contract} that the entire downstream pipeline must satisfy. Second, HyDRA establishes a \textbf{verifiable lineage} by embedding CQ identifiers within the generated ontology's metadata. This unique feature allows any schema component to be programmatically traced back to the specific requirement that motivated its existence, providing an unprecedented level of transparency and auditability.

\subsection{Ontology and Knowledge-Graph Evaluation}

The limitations of traditional static KG evaluation metrics are well documented \cite{brank2005survey}, leading to a call for more functional, task-oriented evaluation methodologies \cite{keet2022evaluate}. Modern approaches have begun to address this by evaluating KGs based on their utility in downstream tasks (\textit{KGrEaT} \cite{ribeiro2021kgreat}) or by developing methods to translate natural language questions into formal queries (\textit{GraphRAFT} \cite{daza2024graphraft}).

HyDRA's evaluation framework represents the logical conclusion of this trend, creating a perfectly symmetrical and closed-loop process. The very CQs that serve as the initial requirement contract are programmatically translated into executable queries for the final artifact. 
%Specifically, we ingest the generated graph into a \textit{Neo4j} database and convert the CQs into \textit{Cypher} queries. 
Specifically, we query the generated graph via verifiable operators using neurosymbolic design principles.
The success rate of these queries provides a direct measure of the KG's functional correctness. This approach ensures that the KG is not just structurally valid but useful for the exact purpose it was designed to serve.

%%%%%%%%%%%%%%%%%%%%%%%%%%%%%%%%%%%%%%%%%%%%%%%%%%%%%%%%%%%%%%%%%%%%%%%%
% ... Other sections of the paper will go here ...
\section{Background}

\subsection{Design-by-Contract}

Design-by-Contract (DbC) is a software engineering methodology introduced by Bertrand Meyer \cite{meyer1992applying, meyer1997object} that emphasizes formal specifications for software components. DbC treats the relationship between a component and its clients as a formal agreement, similar to a legal contract, where each party has obligations and benefits. This methodology is built on three key elements: preconditions (obligations on the caller), postconditions (obligations on the component), and invariants (conditions that must always hold). DbC enhances software reliability by making explicit the assumptions and guarantees of each component, enabling systematic verification and validation throughout the development lifecycle. 
This paradigm offers the theoretical blueprint for HyDRA. 

In recent years, researchers have begun adapting these concepts to AI systems. This has led to probabilistic contracts that can handle model uncertainty \cite{mohajerani2024theory}, assume-guarantee reasoning frameworks for compositional verification like \textit{Pacti} \cite{bian2023pacti}, and runtime enforcement tools like \textit{MLGuard} \cite{ghosh2023mlguard} and \textit{NVIDIA NeMo Guardrails} \cite{nvidia2023nemo} that act as safety-oriented wrappers for ML/LLM systems.

HyDRA adapts these DbC principles to KG construction, providing a framework for ensuring correctness, consistency, and completeness throughout the KG development process. By formalizing the requirements and expectations at each stage, DbC principles enable HyDRA to establish verifiable pathways from conceptual models to implemented knowledge structures. 

\subsection{Definition of Contracts}

A contract in this context comprises four key components: preconditions, postconditions, invariants, and a verification function. Preconditions and postconditions are anologous to the general definition, and invariants establish properties that remain constant throughout the operation. The verification function determines whether these conditions are satisfied and is realized via neurosymbolic operations.

These contracts form a composition of operations in HyDRA, where the postcondition of one phase becomes the precondition of the next \cite{hoare1969axiomatic, lamport1977proving}. This creates a verifiable pathway from initial requirements to final implementation, ensuring traceability and correctness at each step of the KG construction process. When a contract's conditions are violated, HyDRA's verification engine triggers remediation actions to restore compliance, establishing a self-correcting mechanism that maintains the integrity of the KG throughout its development lifecycle.

\subsection{SymbolicAI Framework}

The SymbolicAI framework \citep{dinu2024symbolicai} provides the theoretical and technical foundation for HyDRA's neurosymbolic approach. At its core, SymbolicAI treats LLMs as semantic parsers. The framework introduces polymorphic, compositional, and self-referential operations for multi-modal data that connects multi-step generative processes. SymbolicAI implements a contract-based validation layer that enforces preconditions, postconditions, and type validation using structured data models, with automatic remediation through validation functions and retry logic. This allows HyDRA to take advantage of the pattern recognition capabilities of neural networks (e.g., LLMs) while maintaining the explainability and verifiability of symbolic layers (e.g., contracts). 

\section{HyDRA Architecture}
HyDRA provides an end-to-end framework for building domain-specific ontologies and KGs through an iterative, simulated stakeholder-driven process. HyDRA's architecture is illustrated in Figure~\ref{fig:architecture}. The process begins with a user-defined domain description and some input documents that are used to populate the KG. First, stakeholder groups are generated from the domain description, followed by persona generation for each group. These groups are used to produce so-called scope documents which define the conceptual boundaries of the domain. The scope documents are used to generate competency questions which guide the iterative construction of the ontology. Finally, the resulting ontology and the input documents are used to generate a KG.

Each phase builds on the outputs of the previous steps, ensuring that the final KG is relevant and aligned with the domain. HyDRA does not rely on the input documents for ontology generation, thereby avoiding fitting to the initial, potentially limited sample. Instead, it aims to capture a more general representation of the target domain.

\begin{figure}
    \centering
    \includegraphics[width=1\linewidth]{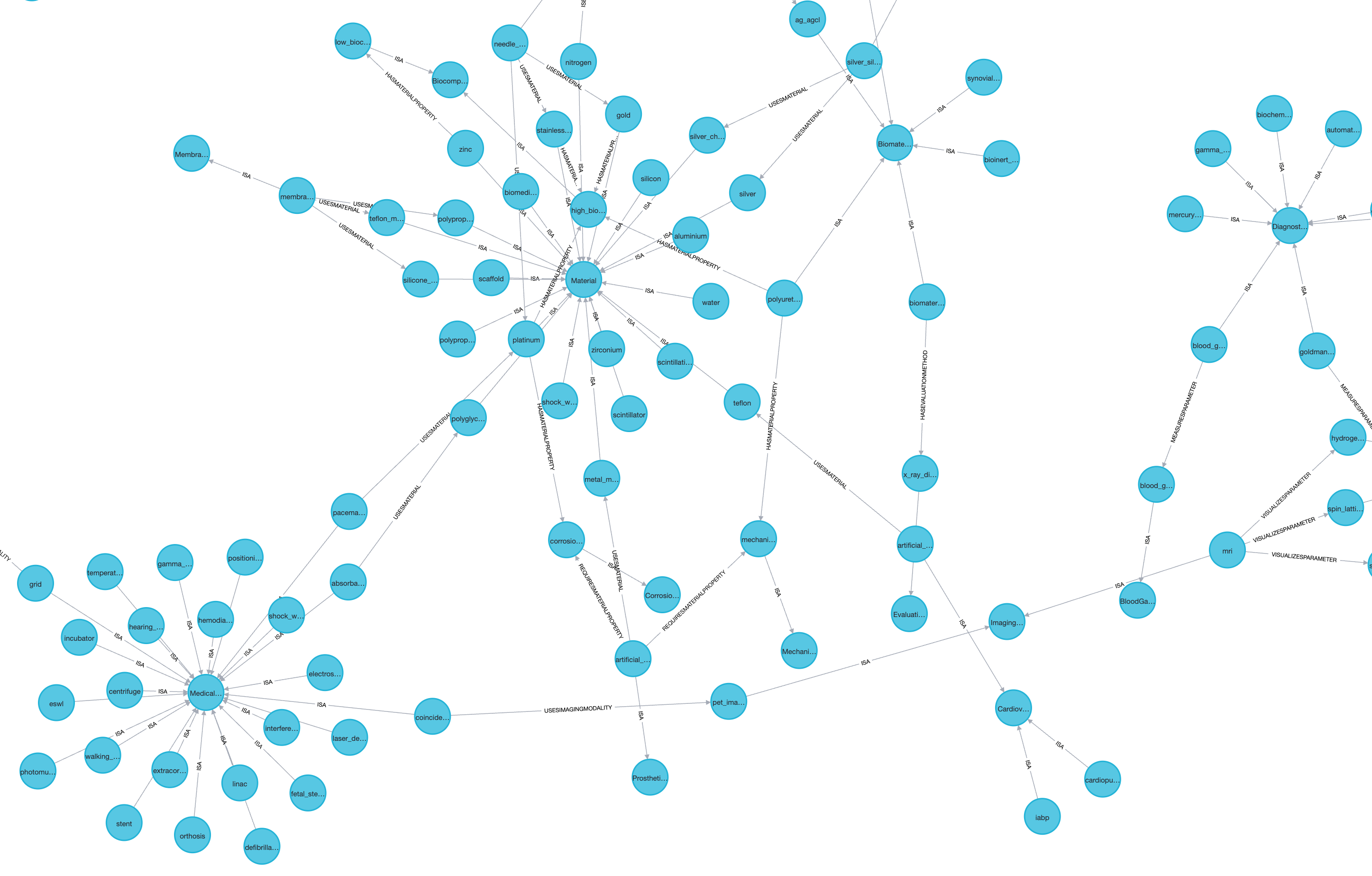}
    \caption{HyDRA-generated, ontology-conditioned KG inserted into Neo4j as one-to-one mapping (partial graph is pictured).}
    \label{fig:kg}
\end{figure}

\subsection{Persona Generation}
Instead of directly generating personas, we first prompt the LLM to produce a prioritized list of stakeholder groups relevant to the domain. For each group, distinct personas are generated, with the number of personas per group increasing with the group's priority. This approach ensures that more influential or central groups are represented by a broader set of perspectives. After generation, all personas are aggregated, shuffled, and divided into $n$ heterogeneous groups $G_i$, where $n$ is set as the number of aggregated personas divided by group size $l$ (with $l=4$ as default hyperparameter). These persona groups are then used as inputs for scope document generation in the next step.

\subsection{Scope Generation}
Given an initial natural-language domain description $d$ provided by the user and the set of heterogenous persona groups identified in the previous phase, HyDRA generates scope documents that define both the conceptual boundaries and the core area of the domain. These scope documents subsequently serve as references for competency question generation. For each persona group $G_i$, one scope document $S_i$ is generated using the LLM and $d$, resulting in an initial set of $n$ scope documents. The scope documents are then recursively merged in batches of size $n$ (with $k=6$ as the default hyperparameter) until a single, consolidated scope document remains. The complete process is outlined in Algorithm~\ref{alg:scope_generation}.

\begin{algorithm}
\caption{Scope Document Generation}
\label{alg:scope_generation}
\begin{algorithmic}[1]
\State \textbf{Input:} Domain description \( d \), persona groups \( \mathcal{G} = \{G_1, G_2, \ldots, G_n\} \), batch size \( k \)
\State \( \mathcal{S} \gets \emptyset \)
\For{each persona group \( G_i \) in \( \mathcal{G} \)}
    \State \( S_i \gets \text{LLM}(d, G_i, \text{prompt}_{\text{scope}}) \)
    \State \( \mathcal{S} \gets \mathcal{S} \cup \{S_i\} \)
\EndFor

\While{\( |\mathcal{S}| > 1 \)}
    \State Partition \( \mathcal{S} \) into batches of size \( k \) (default \( k=6 \))
    \State \( \mathcal{S}_{\text{new}} \gets \emptyset \)
    \For{each batch \( B \) in partition}
        \State \( S_{\text{merged}} \gets \text{LLM}(d, B, \text{prompt}_{\text{merge}}) \)
        \State \( \mathcal{S}_{\text{new}} \gets \mathcal{S}_{\text{new}} \cup \{S_{\text{merged}}\} \)
    \EndFor
    \State \( \mathcal{S} \gets \mathcal{S}_{\text{new}} \)
\EndWhile
\State \textbf{return} the final scope document in \( \mathcal{S} \)
\end{algorithmic}
\end{algorithm}

\subsection{Competency Question Generation}
Taking the previously generated persona groups $G_i$ and the final scope document, HyDRA generates $n$ batches of competency questions that are then merged and deduplicated via a simple LLM-based aggregation step. Competency questions are natural language queries are used to define the requirements and intended capabilities of an ontology. The competency questions are then passed to the ontology generator.

\subsection{Ontology Generation}
For simplicity, we focus on a subset of features defined in OWL2---a standardized ontology language for fomally representing classes, properties and relationships. The ontology generation phase of HyDRA can define class hierarchies, data properties, and object properties, along with characteristics such as \textit{functional} properties (which uniquely associate an entity with a value) or \textit{inverse} properties (which define bidirectional relationships between entities). HyDRA iteratively builds the ontology based on batches of competency questions. Algorithm~\ref{alg:ontology_generation} outlines this process. The ontology $\mathcal{O}$ is updated in each iteration with new concepts $C_i$ generated by the LLM for a batch of competency questions $Q_i$. The candidate concepts $C_i$ are merged with the current ontology to form a candidate ontology $\mathcal{O}'$. If $\mathcal{O}'$ does not violate any constraints, it becomes the new ontology; otherwise, the LLM is repeatedly called to revise its output until all constraints are satisfied. The process uses two types of prompts: $\text{prompt}_{\text{ontology}}$ for initial ontology construction and $\text{prompt}_{\text{fix}_\text{o}}$ for revision in the case of constraint violations.

\begin{algorithm}
\caption{Ontology Generation Algorithm}
\label{alg:ontology_generation}
\begin{algorithmic}[1]
 \State \textbf{Input:} A set of competency questions, partitioned into batches \( \mathcal{Q} = \{Q_1, Q_2, \ldots, Q_m\} \)
\State $\mathcal{O} \gets \emptyset$
\For{each batch of competency questions $Q_i$}
    \State $C_i \gets \text{LLM}(Q_i, \mathcal{O}, \text{prompt}_{\text{ontology}})$ 
    \State $\mathcal{O}' \gets \mathcal{O} \cup C_i$
    \State $\text{Violations} \gets \text{FindViolations}(\mathcal{O}')$
    \While{$\text{Violations} \neq \emptyset$}
        \State $C_i \gets \text{LLM}(\text{Violations}, C_i, \text{prompt}_{\text{fix}_\text{o}})$ 
        \State $\mathcal{O}' \gets \mathcal{O} \cup C_i$
        \State $\text{Violations} \gets \text{FindViolations}(\mathcal{O}')$
    \EndWhile
    \State $\mathcal{O} \gets \mathcal{O}'$ 
\EndFor
\State \textbf{return} $\mathcal{O}$
\end{algorithmic}
\end{algorithm}

\subsubsection{Validation}
The validation step ensures that each candidate ontology produced during generation satisfies a set of predefined constraints before being accepted. HyDRA applies the contract pattern to enforce these constraints after each merge of new concepts. Constraints are checked with respect to the current ontology. Thus, as the ontology grows, validation effectively becomes stricter. For clarity, we distinguish between simple constraints and complex constraints:

\paragraph{Simple Constraints} Simple constraints are fundamental syntactic and referential checks that validate individual ontology elements in isolation. These include naming conventions (class names must start with uppercase letters), uniqueness checks (no duplicate classes or properties), existence validations (referenced classes must exist), and property domain/range validations (ensuring properties reference valid classes).

\paragraph{Complex Constraints} Complex constraints are structural validations that ensure the ontology maintains a proper hierarchical structure. These include preventing circular inheritance relationships (where class A extends B while B extends A directly or indirectly), prohibiting self-inheritance (a class cannot be its own superclass), enforcing single inheritance (each class should have exactly one superclass except the root), and maintaining a single root class (preventing multiple disconnected hierarchies).

\paragraph{Retries} Whenever at least one constraint is violated during generation, the LLM is prompted with both the violations and its previous output to revise its proposal. HyDRA only accepts and merges concepts into the ontology if no violations occur, thus ensuring correctness at each step. Accepting only the non-violating subset and prompting the model to fix only the violations would restrict the LLM from restructuring its output, including changes or removals of concepts that did not initially violate constraints.

\subsection{Knowledge Graph Generation}
HyDRA uses the generated ontology to populate a KG from input data such as text documents. KG generation, like ontology generation, is performed iteratively and additively. HyDRA splits the input documents into chunks and extracts candidate triplets of the form (subject, predicate, object), with a special predicate \textit{isA} for class assignments. Triplet extraction is performed over multiple epochs to allow information from different document parts to be connected. As opposed to ontology generation (where the whole candidate ontology is validated at each step), only the newly-generated triplets are validated. In Algorithm~\ref{alg:kg_generation}, the KG $\mathcal{K}$ is incrementally updated with candidate triplets $T_j$ generated by the LLM for each document chunk $D_j$. The ontology $\mathcal{O}$ provides type and property constraints for validation. Two types of prompts are used: $\text{prompt}_{\text{kg}}$ for initial triplet extraction and $\text{prompt}_{\text{fix}_\text{kg}}$ for revision in case of violations.

\begin{algorithm}
\caption{Knowledge Graph Generation Algorithm}
\label{alg:kg_generation}
\begin{algorithmic}[1]
 \State \textbf{Input:} A set of document chunks \( \mathcal{D} = \{D_1, D_2, \ldots, D_p\} \), an ontology \( \mathcal{O} \)
\State $\mathcal{K} \gets \emptyset$
\For{each epoch}
    \For{each document chunk $D_j$}
        \State $T_j \gets \text{LLM}(D_j, \mathcal{K}, \mathcal{O}, \text{prompt}_{\text{kg}})$
        \State $\text{Violations} \gets \text{FindViolations}(T_j, \mathcal{K}, \mathcal{O})$
        \While{$\text{Violations} \neq \emptyset$}
            \State $T_j \gets \text{LLM}(\text{Violations}, T_j, \text{prompt}_{\text{fix}_\text{kg}})$
            \State $\text{Violations} \gets \text{FindViolations}(T_j, \mathcal{K}, \mathcal{O})$
        \EndWhile
        \State $\mathcal{K} \gets \mathcal{K} \cup T_j$
    \EndFor
\EndFor
\State \textbf{return} $\mathcal{K}$
\end{algorithmic}
\end{algorithm}

\subsubsection{Triplet Validation}
Triplet validation ensures that each candidate KG remains syntactically and semantically consistent with the ontology and all previously accepted triplets. For each triplet, HyDRA applies a set of simple constraints: checking for duplicate entities, enforcing naming conventions, and verifying that properties and class assignments are valid according to the ontology. 

\paragraph{Class Reassignment}
While HyDRA generally disallows reassigning an entity to a completely different class, it does permit class narrowing. If an entity is initially assigned to class $A$, and a later triplet assigns it to a subclass $B$ of $A$, this reassignment is accepted since $B$ inherits all properties from $A$. This reflects the assumption that different documents may progressively specify entity types with increasing precision.

\section{Experimental Setup}

Assessing the quality of a KG, and by proxy the ontology, is difficult as, to the best of our knowledge, no benchmarks exist for this purpose.
Therefore, we turn to the task of question-answering for initial experiments as benchmarks are readily available and KGs are well suited for this task.
While we identified several issues with this evaluation we find our results interesting as they help us to design a novel evaluation scheme, 

\subsection{Benchmark Configuration}

We used the Biomedical Engineering subset of the MedExQA dataset~\cite{kim2024medexqa} for evaluation due to its robust question-answer structure. 
The MedExQA dataset was specifically chosen for its strength in capturing structured question-answer pairs relevant across multiple domains in the medical field. While not part of this work this enables the iterative construction of a KG spanning the full domain.
In this work, we explore HyDRAs capabilities on the biomedical engineering domain of MedExQA.
%Our results are  pick the \textit{Biomedical Engineering} subset from this dataset 
%Although initially planned, the additional MedExQA subsets—Clinical Laboratory Science, Clinical Psychology, Occupational Therapy, and Speech Pathology—were not implemented due to time constraints.

\subsection{Benchmark Domain}

We selected the Biomedical Engineering domain from the MedExQA dataset to assess HyDRA effectiveness and generalizability. The dataset contains four additional domains of interest but we restricted our experiments to a single domain due to computational constraints.

\begin{itemize}
\item \textbf{Biomedical Engineering (MedExQA)}: Representing biomedical devices, methodologies, diagnostic and therapeutic procedures, engineering principles, and patient care interactions.
\end{itemize}

For the benchmark, we created a concise domain-specific description derived from the Biomedical Engineering subset of MedExQA to generate an ontology. This ontology, combined with the structured question-answer data, was then used to construct a KG. We compare this approach to a baseline KG constructed solely from the structured question-answer dataset without any domain description or ontology generation. Both KGs were subsequently evaluated using OpenAI's GPT-4.1 \cite{openai2025gpt41} to assess their effectiveness

\subsection{Evaluation Framework}

We compared our ontology-enhanced KG construction approach against a baseline method utilizing only the structured question-answer data from MedExQA, without domain descriptions or ontology generation. Evaluation of both approaches was conducted using GPT-4.1. All experiments were repeated three times with different random seeds to account for the variability in the LLM output. We report the averages.

We defined and implemented a contract that takes as input the KG, the question, and the possible answers, and outputs a best guess based solely on the provided KG. 
A fuzzy matching algorithm is then used to detect whether an answer is correct based on the expected answer provided by the dataset.
%TODO @Ryan add more details. it should be feasible to reproduce the evaluation from the text here.

\section{Results and Discussion}

\subsection{Quantitative Results}

Our evaluation shows that the KG constructed without an ontology achieved higher accuracy compared to the ontology-enhanced approach. Our quantitative experimental results for the 143 question-answer pairs of the Biomedical Engineering MedExQA Dataset are summarized below:

\begin{table}[h!]
\centering
\begin{tabular}{lccc}
\toprule
 & \textbf{Gemini-2.5-Flash} & \textbf{o4-mini} & \textbf{o3} \\
\midrule
Ontology \\
Runtime          & 17.39  & 48.86   & 214.04 \\
KG w/ ontology \\
Runtime        & 65.35  & 63.41   & 98.10 \\
KG w/o ontology \\
Runtime       & 34.06  & 35.68   & 69.87 \\
Evaluation \\
Runtime              & 8.14   & 6.77    & 7.89 \\
\midrule
HyDRA                 & \textbf{57.3\%}    & \textbf{42.0\%}   & \textbf{61.5\%} \\
Baseline                & \textbf{95.1\%}   & \textbf{97.2\%}  & \textbf{97.9\%} \\
\bottomrule
\end{tabular}
\label{tab:performance}
\caption{Performance (accuracy) and runtime metrics (in minutes) for KG construction and evaluation with and without ontology across different models. }
\end{table}

\subsection{Analysis of Benchmark-Induced Artifacts}

The results presented in Table~\ref{tab:performance} reveal a counterintuitive outcome: the unconstrained model achieves higher performance than its ontology-guided counterpart. A  qualitative analysis of the evaluation data suggests that this is a consequence of the benchmark's inherent simplicity rather than a failure of the ontology-guided method.

We found that most queries in the benchmark are solvable via a single relational hop. Such relations are directly extracted as triplets during the construction of the unconstrained KG. Because this process connects nodes from the document corpora without strict semantic constraints, the LLM can often answer these queries through a straightforward retrieval mechanism. This evaluation scheme, therefore, fails to test for the very scenarios where an ontology provides a distinct advantage: tasks that demand multi-hop inference to construct complex dependencies between entities. A benchmark with a reliable methodology for assessing such multi-step reasoning would be required to demonstrate the benefits of an ontological structure.

\begin{figure}[h]
    \centering
    \includegraphics[width=1\linewidth]{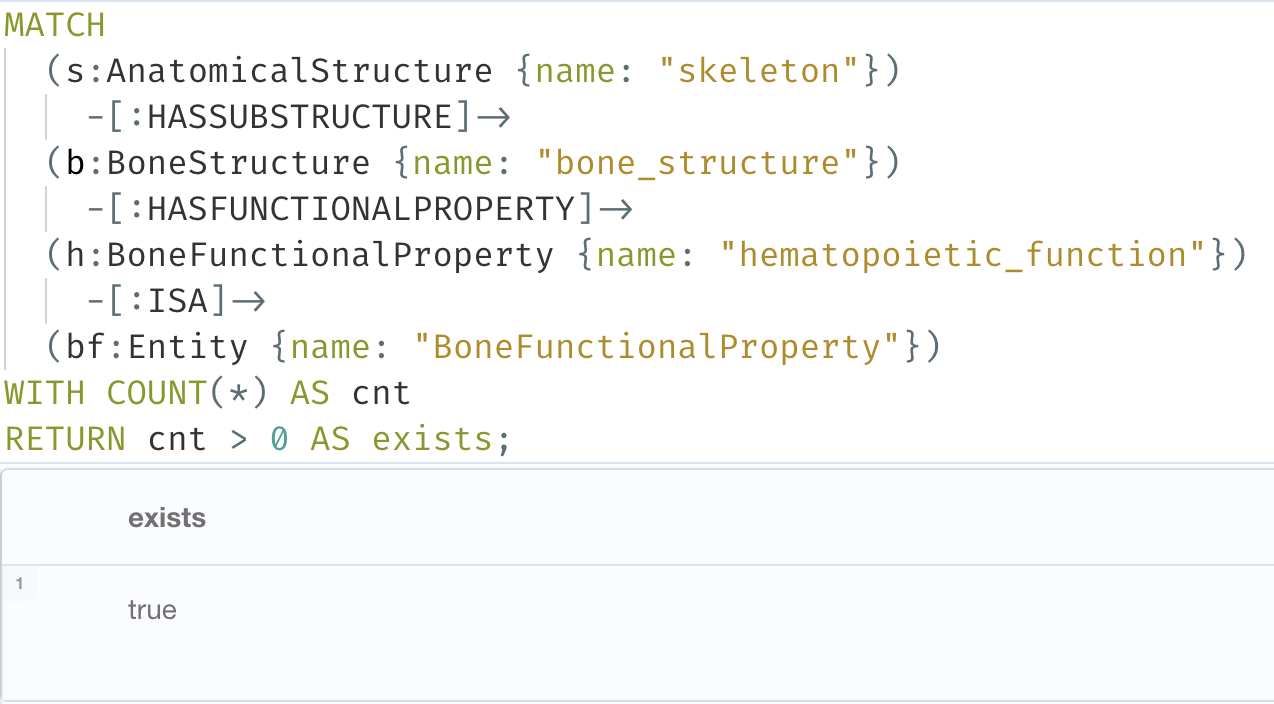}
    \caption{We use the generated Neo4j Cypher query to directly query the knowledge graph that has been inserted into a Neo4j database. The chain of relationships exists, so we obtain a boolean \texttt{true} value. This Cypher query translates to the natural language question: ``Is the skeleton composed of a bone structure that has the functional property \textit{hematopoietic\_function}, and is that \textit{hematopoietic\_function} classified as a \textit{BoneFunctionalProperty}?''}
    \label{fig:results_cypher}
\end{figure}

To illustrate this, we generate 3 prepositional logic statements in both natural language and converted to Neo4j cypher queries. Answering these statements requires at least 2 or 3 node traversals.

\begin{itemize}
    \item Is the skeleton composed of a bone structure that has the functional property \textit{hematopoietic\_function}, and is that \textit{hematopoietic\_function} classified as a \textit{BoneFunctionalProperty}?
    \item Does the patient monitoring device measure body temperature, and is that body temperature classified as a physiological parameter?
    \item Does the artificial blood vessel use teflon as a material, and is teflon classified as a Material?
\end{itemize}

\begin{figure}[ht]
    \centering
    \includegraphics[width=0.8\linewidth]{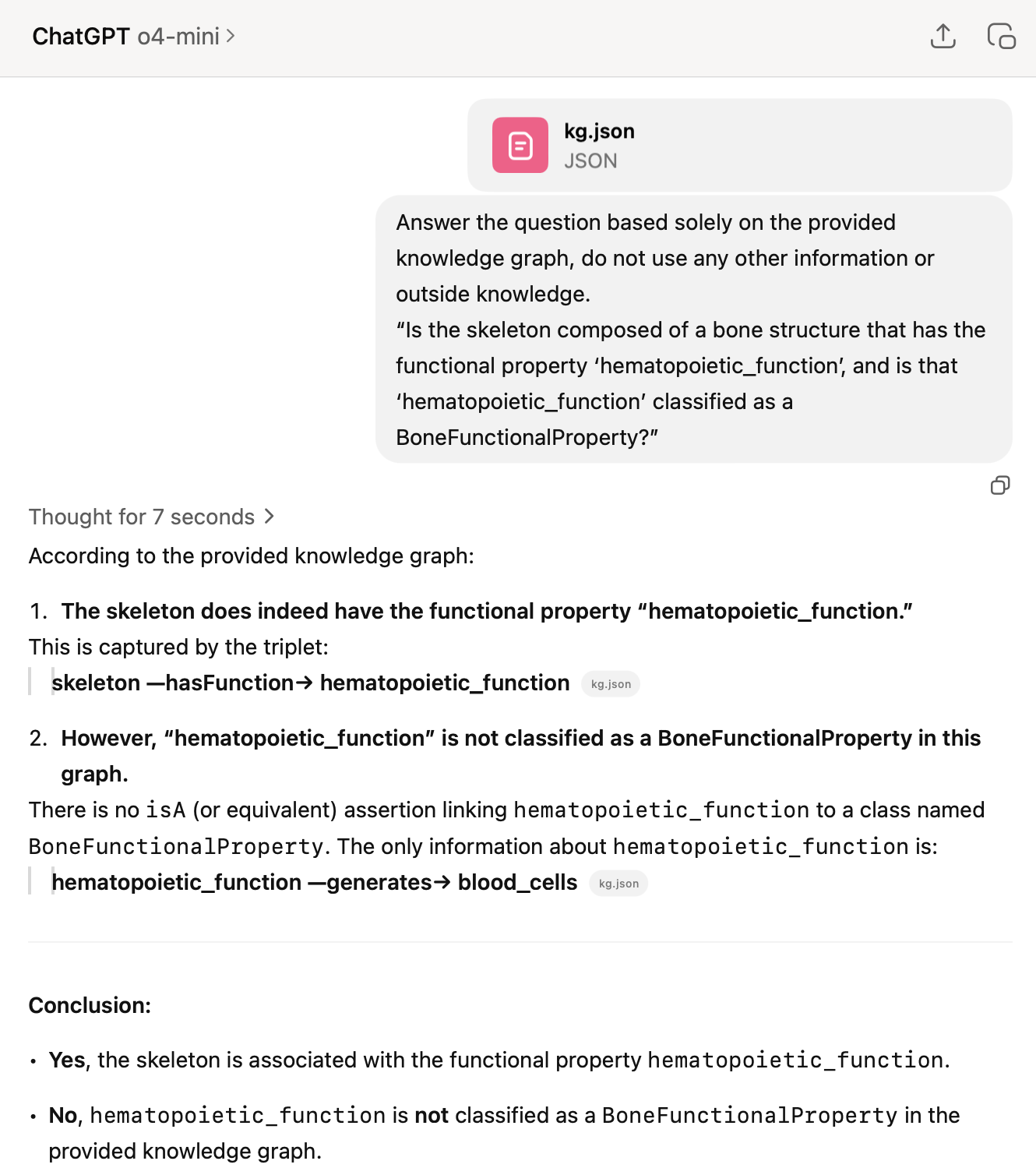}
    \caption{We prompt OpenAI's o4-mini with the natural language version of the cypher query, given the same knowledge graph. The model answers that the first relationship exists, but cannot validate the second condition.}
    \label{fig:results_llm}
\end{figure}

Figure \ref{fig:results_cypher} shows one of these cypher queries and its results as an example. The query can only be resolved by following several edges through the knowledge graph.
Figure \ref{fig:results_llm} shows the results of using the same knowledge graph provided as context to an LLM and prompted with the natural language version of the query. 
All three questions can be correctly answered by simple cypher queries but LLMs fail to correctly answer these statements.

\subsection{Discussion}\label{sec:results_discussion}

These results highlight a clear trade-off between flexibility and exactness, and structural consistency. The higher accuracy observed in the non-ontology approach can be attributed to greater freedom and exactness in generating subject, predicate, object triplets from the question-answer data. Conversely, the ontology-based method intentionally restricts the knowledge representation to maintain consistency and shared structure. This becomes critical when integrating knowledge from multiple diverse sources and doing higher level reasoning on KGs, which isn’t necessarily represented by the questions in the MedExQA dataset. The structured nature provided by the ontology ensures greater compatibility and consistency across various knowledge sources, but restricts the amount of nodes and relationships generated and the exactness in regards to the original data. 

We believe that the observed performance gap is mainly caused by the restrictive nature of HyDRA's ontology. This effect is especially prominent in question-answering benchmarks when the questions can be answered by identifying the information of a single node, as is often the case.
However, once we add iterative KG construction and more complex queries, we believe that our approach will perform much better than the baseline.

Additionally, we identified the LLM-based evaluation as problematic, as it is impossible to distinguish between the models knowledge/capabilities and the quality of the extracted graph.
Therefore, we propose a functional evaluation using graph databases such as Neo4j.
Instead of providing the KG as context to an LLM, it is stored in a graph database.
Then, the questions of the benchmark dataset are automatically translated into graph database queries.
This evaluation has two main advantages:
\begin{enumerate}
    \item The quality of the graph is evaluated directly.
    \item The complexity of the queries can be adjusted in a controlled manner to test for reasoning over many nodes or simple knowledge retrieval.
\end{enumerate}
However, this evaluation scheme is still a work in progress % due to extensive processing times and the significant expansion of context space required by the LLM model when handling large graphs. 
as it proved difficult to automatically generate accurate graph database queries suitable for reliable evaluation. This is a problem we are confident can be solved. %Importantly, the HyDRA-generated KG can be loaded into Neo4j \emph{as-is} (Figure~\ref{fig:kg}): no structural remapping or loss of information is required, so the perceived difficulty of "porting" a KG to Neo4j is confined to query formulation, not ingestion.
Importantly, HyDRA-generated KGs can be loaded into Neo4j as-is (Figure~\ref{fig:kg}). No structural remapping or information reduction is required, so the perceived difficulty of integrating KGs with Neo4j is limited by the generative capabilities of LLMs for query formulation, rather than graph ingestion.

\subsection{Limitations and Challenges}

The main limitation and challenge we encountered was the lack of benchmark datasets that are suitable for measuring the quality of a graph based knowledge representation and retrieval systems.
As discussed in section \ref{sec:results_discussion} we believe that the performance numbers we present in table \ref{tab:performance}, while accurate for simple knowledge retrieval tasks, do not represent the full capabilities of HyDRA's ontology based approach.
However, we acknowledge that compiling a benchmark with complex queries that require a well-defined and consistent graph is a difficult task.
But we believe, that this is a worthwhile endeavor which will help in developing and comparing better KG construction methods, ultimately benefiting neurosymbolic applications.

\section{Lessons learned}
\paragraph{Validation} We were able to identify constraints that simplified the ontology generation phase and improved output quality. In an early version of HyDRA, the ontology generation validation was less strict and allowed for multiple class hierarchies. To remediate, we implemented a contract for a subsequent ontology fixing phase to merge the class hierarchies to a single tree using a pre-defined set of operations. Later, we were able to replace this fixing phase by introducing additional (complex) constraints in the generation phase to enforce a strict hierarchy.

We assume that the constraints we identified are not exhaustive and that quality and efficiency can be improved even more by identifying further constraints.

\paragraph{Automatic fixing} During evaluation, we observed a number of mistakes made by the LLMs which could be solved automatically without having to prompt the LLM again. In some instances, the KG generation phase yields triplets where subject and object are swapped (\texttt{(\scriptsize{object, predicate, subject)}} instead of \texttt{\scriptsize{(subject, predicate, object)}}). The resulting constraint violation lead to regeneration of the entire set of triplets output by the LLM. Instead, if such a constraint violation occurred, we now automatically swap object and subject if the resulting triplet would not violate any constraints.

We expect that there are additional opportunities to short-circuit the generation process in similar ways. By handling these cases deterministically, one could reduce uncertainty and dependence on LLM outputs.

\paragraph{Constraint bypasses} In some instances where the generated KG triplets violates many constraints, we observed that the models simply drop all (or most of the) violating triplets instead of trying to fix them. We assume that this behavior could be reduced by either presenting errors incrementally instead of all at once, increasing both time and costs, or by requiring the model to send \textit{delta updates} to modify the initial output accordingly.

\paragraph{Evaluation} The observed performance gap between the ontology-enhanced and baseline approaches likely arises due to inherent biases in the evaluation setup. KG generation without an ontology effectively encodes documents directly into the context with minimal abstraction, thereby maximizing the relevance to the evaluation questions. Conversely, ontology-based KG construction does not directly depend on input documents and thus can fully represent all relevant information only if the ontology itself is infinitely detailed, which is not feasible, or if the competency questions precisely match the evaluation questions, which introduces significant evaluation bias.

Additionally, the chosen dataset primarily evaluates question-answering capabilities based on explicit information from source documents, rather than testing the type of inferential reasoning that ontologies typically support. Therefore, future evaluations should include benchmarks specifically designed to assess the inferential and structural reasoning benefits provided by ontology-driven KG approaches. We outline one such approach in section \ref{sec:results_discussion}

\section{Conclusion}
In this work, we introduced HyDRA, a Hybrid‑Driven Reasoning Architecture that operationalizes neurosymbolic principles and Design‑by‑Contract concepts to guide and verify large language model (LLM)‑driven knowledge graph (KG) construction. Our approach directly addresses key challenges in automated KG generation—most notably entity resolution, structural consistency, and the risk of producing isolated or semantically incoherent graph fragments—by following using verifiable contracts, a neurosymbolic approach following design-by-contract, throughout the construction pipeline.

Through experiments on the MedExQA benchmark, we observed that a baseline, ontology‑free KG construction approach achieved higher raw accuracy on simple single‑hop question answering tasks. While initially counterintuitive, this result reflects a limitation of current benchmarks rather than a fundamental weakness of ontology‑guided methods. Our analysis shows that existing evaluation datasets fail to test the scenarios where ontology guidance offers distinct advantages: multi‑hop reasoning, cross‑document integration, and enforcing global structural consistency.

The HyDRA framework demonstrates that incorporating explicit constraints and iterative repair mechanisms can improve the verifiability and maintainability of automatically generated KGs, even if this comes at the cost of reduced exactness in certain narrow evaluation settings. Importantly, HyDRA opens the door for richer, contract‑driven approaches where symbolic layers continuously steer and refine neural outputs.

Looking ahead, we see three key directions: (1) developing evaluation benchmarks that measure inferential reasoning and structural soundness rather than mere retrieval; (2) exploring additional constraints and deterministic fixes to further reduce LLM errors and runtime costs; and (3) extending HyDRA to support large‑scale, iterative KG construction across heterogeneous domains. By advancing these lines of work, we aim to enable the reliable, scalable deployment of neurosymbolic AI systems that leverage the complementary strengths of neural generation and symbolic verification.

% \begin{ack}
% By using the \texttt{ack} environment to insert your (optional)
% acknowledgements, you can ensure that the text is suppressed whenever
% you use the \texttt{doubleblind} option. In the final version,
% acknowledgements may be included on the extra page intended for references.
% \end{ack}
%%%%%%%%%%%%%%%%%%%%%%%%%%%%%%%%%%%%%%%%%%%%%%%%%%%%%%%%%%%%%%%%%%%%%%%%
\bibliography{mybibfile}
\end{document}